\documentclass[11pt,ruled,lined]{article}
\usepackage[numbers]{natbib}
\usepackage{authblk}
\usepackage[T1]{fontenc}
\usepackage[utf8]{inputenc}
\usepackage{algorithm2e}
\usepackage{amsthm}
\usepackage{amsmath}
\usepackage{graphicx}

\usepackage{geometry}
\makeatletter

\providecommand{\tabularnewline}{\\}

\theoremstyle{plain}
\newtheorem{thm}{\protect\theoremname}

\usepackage{url}
\usepackage{amsfonts}
\usepackage{nicefrac}
\usepackage{microtype}

\title{Simultaneous Sparse Dictionary Learning and Pruning}

\author{Simeng Qu}
\author{Xiao Wang}
\affil{Department of Statistics, Purdue University}

\providecommand{\theoremname}{Theorem}

\makeatother

\begin{document}


\maketitle

\begin{abstract}
Dictionary learning is a cutting-edge area in imaging processing,
that has recently led to state-of-the-art results in many signal processing
tasks. The idea is to conduct a linear decomposition of a signal using
a few atoms of a learned and usually over-completed dictionary instead
of a pre-defined basis. Determining a proper size of the to-be-learned
dictionary is crucial for both precision and efficiency of the process,
while most of the existing dictionary learning algorithms choose the
size quite arbitrarily. In this paper, a novel regularization method
called the Grouped Smoothly Clipped Absolute Deviation (GSCAD) is
employed for learning the dictionary. The proposed method can simultaneously
learn a sparse dictionary and select the appropriate dictionary size.
Efficient algorithm is designed based on the alternative direction
method of multipliers (ADMM) which decomposes the joint non-convex
problem with the non-convex penalty into two convex optimization problems.
Several examples are presented for image denoising and the experimental
results are compared with other state-of-the-art approaches. 
\end{abstract}

\section{Introduction}

Sparse coding which represents a signal as a sparse linear combination
of a representation basis in a dictionary has been successfully applied
in many signal processing tasks, such as image restoration \citep{elad2006,yang2008image},
image classification \citep{wright2009robust,yang2009linear}, to
name a few. The dictionary is crucial and plays an important role
in the success of sparse representation. Most of the compressive sensing
literature takes off-the-shelf bases such as wavelets as the dictionary
\citep{candes2006stable,donoho2006compressed}. In contrast, dictionary
learning assumes that a signal can be sparsely represented in a learned
and usually over-completed dictionary. The pre-specified dictionary
might be universal but will not be effective enough for specific task
such as face recognition\citep{zhang2010discriminative,kong2012dictionary}.
Instead, using the learned dictionary has recently led to state-of-the-art
results in many practical applications, such as denoising \citep{elad2006,mairal2012task,zhou2012nonparametric,aharon2006},
inpainting \citep{mairal2009online,mairal2009supervised,Ranzato2006efficient},
and image compression \citep{bryt2008compression}.

In this paper, we propose a novel regularization method called the
Grouped Smoothly Clipped Absolute Deviation (GSCAD) to learn a sparse
dictionary and select the appropriate dictionary size simultaneously. 
It should be emphasized that determining a proper size of the to-be-learned
dictionary is crucial for both precision and efficiency of the process.
There are not too many existing work on discussing the selection of
the dictionary size while most algorithms fix the number of atoms
in the dictionary. In general, a two-stage procedure may be used to
infer the dictionary size by first learning a dictionary with a fixed
size then defining a new objective function penalizing the model complexity
\citep{gassiat2013}. The Bayesian technique can be also employed
by putting a prior on the dictionary size \citep{zhou2009}. In addition, many methods have addressed the group variable selection problem in statistics literature \citep{yuan2006model,zhao2006grouped, huang2009group,zhou2010group,geng2015group}.

This paper makes four main contributions: 
\begin{itemize}
\item Our approach imposes sparsity-enforcing constraints on the learned
atoms, which improves interpretability of the results and achieves
variable selection in the input space. 
\item Our approach is a one-stage procedure to learn a sparse dictionary
and the dictionary size jointly. 
\item Our proposed algorithm is based on the alternative direction method
of multipliers (ADMM) \citep{boyd2011}. The joint non-convex problem
with the non-convex penalty is decomposed into two convex optimization
problems. 
\item Compared with other state-of-the-art dictionary learning methods,
GSCAD has better or competitive performance in various denoising tasks. 
\end{itemize}

\section{GSCAD penalty}

\textbf{Review of the Smoothly Clipped Absolute Deviation (SCAD) penalty}.
SCAD penalty is first proposed by \citep{fan2001} in the context
of high dimensional linear regression. SCAD has some desired properties:
(i) Unbiasedness: the resulting estimator is nearly unbiased when
the true unknown parameter is large; (ii) Sparsity: The resulting
estimator is able to sets small estimated coefficients to zero to
reduce model complexity; (iii) Continuity: The resulting estimator
is continuous in data to avoid instability in model prediction. Defined
as

\begin{equation}
\psi_{\lambda}(d)=\begin{cases}
\lambda|d|, & if\ |d|\leq\lambda\\
-\frac{|d|^{2}-2c\lambda|d|+\lambda^{2}}{2(c-1)}, & if\ \lambda<|d|\leq c\lambda\\
\frac{(c+1)\lambda^{2}}{2}, & if\ |d|>c\lambda
\end{cases},\label{eq:SCAD}
\end{equation}
for some $\lambda>0$ and $c>2$, the SCAD contains three segments.
When $d$ is small (less than $\lambda$), it acts exactly like the
Lasso penalty; when $d$ is big (greater than $3\lambda$), it becomes
a constant so that no extra penalty is applied to truly significant
parameters; these two segments are connected by a quadratic function
which results in a continuous differentiable SCAD penalty function
$\psi_{\lambda}(\cdot)$.

\textbf{GSCAD penalty}. Even though the SCAD penalty possesses many
good properties, it only treats parameters individually and does not
address any group effect among parameters. With respect to the structure
of the dictionary, we propose a new penalty, GSCAD, where G stands
for group. Let $\theta$ be a vector in $\mathbb{R}^{m}$. The GSCAD
penalty is defined as 
\[
\Psi_{\lambda}(\theta)=\log\{1+\sum_{k=1}^{m}\psi_{\lambda}(\theta_{k})\},
\]
where $\psi_{\lambda}$ is the SCAD penalty defined in ($\ref{eq:SCAD}$).
It inherits all three merits of SCAD, unbiasedness, sparsity and continuity,
and at the same time takes into account both individual parameters
and group effect among parameters. Individually, the GSCAD penalty
tends to set small estimated $\theta_{k}$ to zero. Group-wise, if
all elements in $\theta$ are small, the penalty will penalize the
entire vector $\theta$ to zero. In addition, if some of the $\theta_{k}$
is significantly large, the penalty will have more tolerance of smaller
elements appearing in $\theta$.

\begin{figure}
\includegraphics[width=5cm,height=4cm]{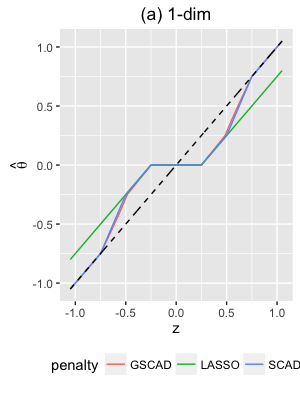}\includegraphics[width=9cm,height=4cm]{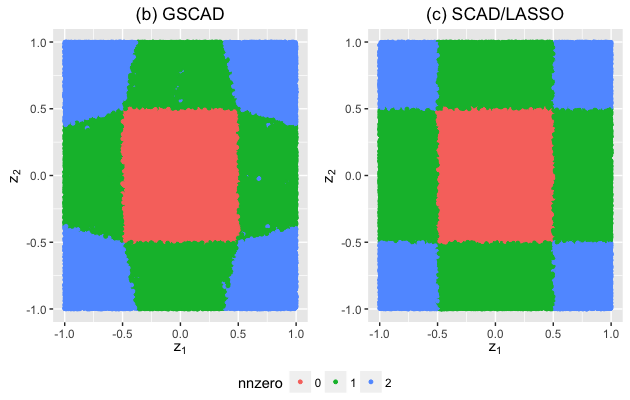}

\protect\caption{\label{fig:penalty}(a) 1-dim threshold function. (b)-(c) Partitions
of the 2-dim space $(z_{1},z_{2})\in\mathbb{R}^{2}$ according to
the number of nonzero elements in $\hat{\theta}$. }
\end{figure}

To better understand GSCAD, let us consider a penalized least squares
problem with an orthogonal design 
\[
\frac{1}{2}\|z-\theta\|_{2}^{2}+p_{\lambda}(|\theta|),
\]
where $z$ and $\theta$ are vectors in $\mathbb{R}^{m}$. For GSCAD,
SCAD and LASSO, the penalty $p_{\lambda}(|\theta|)$ is, respectively,
\[
p_{\lambda}(|\theta|)=\log\{1+\sum_{k=1}^{m}\psi_{\lambda}(\theta_{k})\},~\quad p_{\lambda}(|\theta|)=\sum_{k=1}^{m}\psi_{\lambda}(\theta_{k}),\quad~p_{\lambda}(|\theta|)=\sum_{k=1}^{m}|\theta_{k}|.
\]
Estimators of $\theta$ when $m=1$ are shown in Figure $\ref{fig:penalty}$
(a), where GSCAD performs very similar to SCAD. All three penalties
shows sparsity properties since they all set $\hat{\theta}$ to zero
when $|z|\leq\lambda$. While the soft-thresholding from LASSO has
the inherent bias issue, SCAD and GSCAD give $\hat{\theta}=z$ when
$|z|\geq c\lambda$ and and avoid bias. In a two-dimensional case
when $m=2$ and $z=(z_{1},z_{2})$, we investigate partitions of the
space according to the number of non-zero element in the resulting
estimator $\hat{\theta}=(\hat{\theta}_{1},\hat{\theta}_{2})$, see
Figure $\ref{fig:penalty}$ (b)-(c). While SCAD and Lasso treat each
coordinate individually, GSCAD takes into account the whole group.
It is less likely to set the estimator of one coordinate to zero as
the estimator of another coordinate gets away from zero.

\textbf{Convexity.} Even though GSCAD is build upon the non-convex
penalty function SCAD, our development uncovers a surprising fact
that the optimization problem of GSCAD under orthogonal design is
a convex problem. This will greatly facilitates the implementation
of GSCAD. 
\begin{thm}
\label{Thm: convexity}Define $\hat{\theta}=(\hat{\theta}_{1},...,\hat{\theta}_{m})$
as the minima of optimization problem 
\begin{equation}
\min_{\theta\in\mathbb{R}^{m}}\frac{\varrho}{2}\sum_{k=1}^{m}(z_{k}-\theta_{k})^{2}+\log\{1+\sum_{k=1}^{m}\psi_{\lambda}(\theta_{k})\},\quad\mbox{with constant }\varrho>0.\label{eq:P}
\end{equation}
Then, 
\begin{itemize}
\item[(1)] $sign(\hat{\theta}_{k})=sign(z_{k})$, and $|\hat{\theta}_{k}|\leq|z_{k}|$.
Denote $\tilde{K}=\{1\leq k\leq K:\ z_{k}\neq0\}$, and let $\Theta_{k}$
be the open interval between $z_{k}$ and 0. Then problem $(\ref{eq:P})$
is equivalent to 
\begin{equation}
\min_{\theta_{k}\in\Theta_{k}\cup\{0\},k\in\tilde{K}}\frac{\varrho}{2}\sum_{k\in\tilde{K}}(z_{k}-\theta_{k})^{2}+\log\{1+\sum_{k\in\tilde{K}}\psi_{\lambda}(\theta_{k})\}\label{eq:P2}
\end{equation}

\item[(2)] Let $c_{0}=card(\tilde{K})$, be the number of non-zero element in
$z$. If 
\begin{equation}
\lambda^{2}\leq\varrho c_{0}^{-1}\quad and\quad(c-1)\{\varrho(1+\lambda^{2})^{2}-c_{0}\lambda^{2}\}\geq1+\lambda^{2},\label{eq:c_lambda}
\end{equation}
then optimization problem $(\ref{eq:P2})$ is convex, and $\hat{\theta}$
is continuous in data $z$. 
\end{itemize}
\end{thm}
\textbf{Remarks on Theorem \ref{Thm: convexity}}. (i) Adding a constant
$\varrho$ in (\ref{eq:P}) makes the problem more general such that
the convexity result can be directly applied to the algorithms in
Section $\ref{sub:Algorithms}$, where $\varrho$ plays a role of
penalty parameter in the Augmented Lagrangian method. (ii) Condition
(\ref{eq:c_lambda}) can be satisfied easily under a wide range of
circumstances. For instance, in the previous two-dimensional example
with $\varrho=1$, $c_{0}=2$, and $c=3$, Condition (\ref{eq:c_lambda})
will be satisfied as long as $\lambda\leq2^{-1/2}$.

\section{Dictionary Learning with GSCAD}

\subsection{Matrix Factorization Framework}

Dictionary learning problems are commonly specified under the framework
of matrix factorization. Consider a vectorized clean signal $\mathbf{x}\in\mathbb{R}^{m}$
and a dictionary $\mathbf{D}=(\mathbf{d}_{1},...,\mathbf{d}_{p})\in\mathbb{R}^{m\times p}$
, with its $p$ columns referred to as atoms. Sparse representation
theory assumes that signal $\mathbf{x}$ can be well approximated
by a linear combination of a few atoms in $\mathbf{D}$, i.e. 
\[
\mathbf{x}\approx\mathbf{D}\alpha,
\]
where the number of non-zero elements in $\alpha$ is far less than
the number of atoms $m$. In most of the cases, the clean signal $\mathbf{x}$
won't be available, and instead, we will only be able to observe a
noisy signal $\mathbf{y}=\mathbf{x}+\epsilon$, where $\epsilon$
represents noise with mean zero and variance $\sigma^{2}$. Suppose
we have $n$ signals $\mathbf{Y}=(\mathbf{y}_{1},...,\mathbf{y}_{n})\in\mathbb{R}^{m\times n}$,
and we want to retrieve the corresponding clean signals $\mathbf{X}=(\mathbf{x}_{1},...,\mathbf{x}_{n})$.
This can be summarized as a matrix factorization model 
\[
\mathbf{Y}=\mathbf{D}\mathbf{A}+\mathbf{\epsilon},
\]
where $\mathbf{A}=(\alpha_{1},...,\alpha_{m})$. To make
the problem identifiable, we require the dictionary $\mathbf{D}$
belongs to a convex set $\mathcal{\mathcal{D}}$ 
\[
\mathcal{\mathcal{D}}=\{\mathbf{D}\in\mathbb{R}^{m\times p}\ s.t.\ \forall j=1,...,p,\ ||\mathbf{d}_{j}||_{\infty}\leq1\}.
\]

Dictionary learning aims to obtain estimations of dictionary $\hat{\mathbf{D}}$
and sparse coding $\hat{\mathbf{A}}$, and then reconstruct the clean
signal as $\mathbf{\hat{x}=\hat{\mathbf{D}}\hat{A}}$. This is usually
done by minimizing the total squared error: 
\[
\min||\mathbf{Y}-\mathbf{D}\mathbf{A}||_{F}^{2},\quad\text{ subject to additional sparsity constrains on }\alpha,
\]
where $||\cdot||_{F}$ is the Frobenius norm. Constrains such as $||\alpha||_{0}\leq L$
($l_{0}$-penalty ) and $||\alpha||_{1}\leq\lambda$ (Lasso penalty)
for some positive constants $L$ and $\lambda$ are widely adopted
by dictionary learning literature. Experiments have shown that Lasso
penalty provides better results when used for learning the dictionary,
while $l_{0}$ norm should always be used for the final reconstruction
step \citep{Mairal2014}.

\subsection{Regularization on Dictionary}

Compared with sparse coding, regularization on dictionary size is
less studied. Most of the existing methods, such as K-SVD and Online
Learning, estimate the dictionary directly with a fixed dictionary
size. They usually require the size of the dictionary to be specified
before learning, and this will end up with a solution of over completed
dictionary with $p>m$, which may not be very helpful if we want to
better understand the mechanism. In addition, learning a sparse dictionary
can lower the model complexity and improve interpretability of the
results. All these issues can be addressed with the help of GSCAD
penalty, where we would be able to reveal the real size of the dictionary
and at the same time obtain an estimated sparse dictionary. More specifically,
denote dictionary as $\mathbf{D}$ with $p$ atoms $\mathbf{d}_{i}=(d_{i1},\ldots,d_{im})^{T}\in\mathbb{R}^{m},1\leq i\leq p$.
The GSCAD penalty on dictionary \textbf{$\mathbf{D}$} is defined
by 
\[
\Psi_{\lambda}(\mathbf{D})=\sum_{j=1}^{p}\log\{1+\sum_{k=1}^{m}\psi_{\lambda_{1}}(d_{jk})\}
\]
where $\psi_{\lambda}$ is the SCAD penalty defined in ($\ref{eq:SCAD}$).
The objective function for our problem is formulated as 
\begin{equation}
\min_{\mathbf{D}\in\mathcal{D},\alpha_{i}\in\mathbb{R}^{p}}\frac{1}{2}\sum_{i=1}^{n}||\mathbf{y}_{i}-\mathbf{D}\mathbf{\alpha}_{i}||_{2}^{2}+\Psi_{\lambda_{1}}(\mathbf{D})+\lambda_{2}\sum_{j=1}^{p}||\mathbf{\alpha}_{j}||_{1}.\label{eq:obj}
\end{equation}

Firstly, the GSCAD penalty tends to set small estimated $d_{ij}$
to zero, and reduces the complexity of the estimated dictionary. If
all elements in $\mathbf{d}_{i}$ are small, GSCAD will lead to $\mathbf{d}_{i}=0$.
Therefore, when starting with a relatively large $p$, GSCAD will
be able to prune the dictionary by penalizing useless atoms to zero.
In this way, the true size of the dictionary can be approximated by
the number of non-zero columns in the resulting dictionary. In addition,
if GSCAD detects some significant $d_{ij}$s in\textbf{ $\mathbf{d}_{i}$},
it will exert less penalty on the whole $\mathbf{d}_{i}$ to avoid
mistakenly truncating any real signals.

\subsection{Algorithms \label{sub:Algorithms}}

We follow the classic two steps approach to solve the optimization
problem $(\ref{eq:obj})$ iteratively. Given the dictionary $\mathbf{D}$,
we update $\mathbf{A}=(\alpha_{1},...,\alpha)$ by solving the Lasso
problem, 
\[
\min_{\mathbf{\alpha}_{i}\in\mathbb{R}^{p}}\frac{1}{2}||\mathbf{y}_{i}-\mathbf{D}\mathbf{\alpha}_{i}||_{2}^{2}+\lambda_{2}||\mathbf{\alpha}_{i}||_{1}
\]
for all signals $1\leq i\leq n$. Given $\mathbf{A}$, the optimization
problem $(\ref{eq:obj})$ becomes 
\begin{equation}
\arg\min_{\mathbf{D}\in\mathcal{C}}\frac{1}{2}\sum_{i=1}^{n}||\mathbf{y}_{i}-\mathbf{D}\mathbf{\alpha}_{i}||_{2}^{2}+\Psi_{\lambda_{1}}(\mathbf{D}),\label{eq:up_dic}
\end{equation}
which is addressed by the ADMM algorithm. Once $\mathbf{D}$ is updated,
we remove all zero columns of $\mathbf{D}$ and reset $p$ to the
number of current atoms. Algorithm $\ref{alg:whole}$ demonstrates
this whole procedure. It should be noted that (\ref{eq:up_dic}) is
a non-convex problem. Recently, the global convergence of ADMM in
non-convex optimization is discussed in \citep{wang2015global}, which
shows that several ADMM algorithms including SCAD are guaranteed to
converge.

\textbf{ADMM for updating dictionary.} Problem (\ref{eq:up_dic})
is equivalent to 
\begin{align*}
 & \min\frac{1}{2}\sum_{i=1}^{n}||\mathbf{y}_{i}-\mathbf{D_{1}}\mathbf{\alpha}_{i}||_{2}^{2}+\Psi_{\lambda_{1}}(\mathbf{D_{2}})\\
 & s.t.\quad\mathbf{D_{1}}=\mathbf{D_{2}}.
\end{align*}
We form the augmented Lagrangian as 
\[
L_{\varrho}(\mathbf{D_{1}},\mathbf{D_{2}},\mathbf{\xi})=\frac{1}{2}\sum_{i=1}^{n}||\mathbf{y}_{i}-\mathbf{D_{1}}\mathbf{\alpha}_{i}||_{2}^{2}+\frac{\varrho}{2}||\mathbf{D_{1}}-\mathbf{D_{2}}||_{F}^{2}+\varrho||\xi\circ(\mathbf{D_{1}}-\mathbf{D_{2}})||_{F}+\Psi_{\lambda_{1}}(\mathbf{D_{2}}).
\]
where $\circ$ is the element-wise multiplication operator of two
matrices, and $\xi\in\mathbb{R}^{d\times p}$. The ADMM algorithm
consists three steps in each iteration 
\begin{align}
\mathbf{\mathbf{D_{1}}}^{(t+1)} & =\arg\min_{\mathbf{D_{1}}}L_{\varrho}(\mathbf{D_{1}},\mathbf{D_{2}}^{(t)},\mathbf{\xi}^{(t)})\label{eq:D1}\\
\mathbf{D_{2}}^{(t+1)} & =\arg\min_{\mathbf{D_{2}}}L_{\varrho}(\mathbf{D_{1}}^{(t+1)},\mathbf{D_{2}},\mathbf{\xi}^{(t)})\label{eq:D2}\\
\mathbf{\xi}^{(t+1)} & =\mathbf{\xi}^{(k)}+(\mathbf{D_{1}}^{(k+1)}-\mathbf{D_{2}}^{(k+1)}).\nonumber 
\end{align}
Problem ($\ref{eq:D1}$) bears an explicit solution 
\[
\mathbf{D_{1}}^{(t+1)}\leftarrow\{\frac{1}{m}\mathbf{y}\mathbf{A}^{T}+\varrho(\mathbf{D_{2}}^{(t)}-\mathbf{\mathbf{\xi}}^{(t)})\}(\frac{1}{m}\mathbf{A}\mathbf{A}^{T}+\varrho I_{r})^{-1}.
\]
$\mathbf{D}_{2}$ in ($\ref{eq:D2}$) can be solved by columnwise
optimization such as 
\[
\mathbf{d_{2}}_{j}^{(t+1)}=\arg\min_{\mathbf{d_{2}}_{j}}\frac{\varrho}{2}||\mathbf{d_{2}}_{j}-(\mathbf{d_{1}}_{j}^{(t+1)}+\mathbf{\xi}_{j}^{(t)}||_{2}^{2}+\log\{1+\Psi_{\lambda_{1}}(\mathbf{d_{2}}_{j})\},
\]
for $1\leq j\leq p$. In theorem $\ref{Thm: convexity}$, we have
shown that this is a convex problem under Condition (\ref{eq:c_lambda}),
and can be solved easily by exiting convex optimization algorithms.
The ADMM algorithm for updating dictionaries is summarized in Algorithm
$\ref{alg:Updic}$.

\begin{algorithm}
\SetKwInOut{Input}{Input} 
\SetKwInOut{Output}{Output}
\SetAlgoLined
\Input{Training samples $\mathbf{Y}=[\mathbf{y}_{1},...,\mathbf{y}_{n}]$,
parameter $\lambda_{1}$,$\lambda_{2}$,$c$,$m$,$p_{0}$ }

initialize $\mathbf{D}^{(0)}\in\mathbb{R}^{m\times p_{0}}$ as random
matrix with $d_{ij}\sim Unif(0,1)$\;

\While{not converge}{{Sparse Coding Stage:} for $i=1,...,n$,
update $\mathbf{\alpha}_{i}$ by solving Lasso problem 
\[
\min_{\mathbf{\alpha}_{i}\in\mathbb{R}^{p}}\frac{1}{2}||\mathbf{y}_{i}-\mathbf{D}\mathbf{\alpha}_{i}||_{2}^{2}+\lambda_{2}||\mathbf{\alpha}_{i}||_{1};
\]
{Dictionary Update Stage:} update $\mathbf{D}$ using Algorithm
\ref{alg:Updic}\; Number of atoms: $p\leftarrow\mbox{\# columns of }\mathbf{D}$
}

\Output{$\mathbf{D}$, $p$} \protect\caption{\label{alg:whole}Dictionary Learning with GSCAD}
\end{algorithm}

\begin{algorithm}
\SetKwInOut{Input}{Input} 
\SetKwInOut{Output}{Output} 
\SetAlgoLined

\Input{Training samples $\mathbf{Y}$, current $\mathbf{A}=(\alpha_{1},...,\alpha_{n})$,
parameter $\lambda_{1}$,$c$,$\varrho$} Initialize $\mathbf{D_{2}}^{(0)}=\mathbf{\xi}=\mathbf{0}\in\mathbb{R}^{n\times p}$,
set $t=0$\\
 \While{not converge}{ $\mathbf{D_{1}}^{(t+1)}\leftarrow\{\mathbf{y}\mathbf{A}^{T}+\varrho(\mathbf{D_{2}}^{(t)}-\mathbf{\mathbf{\xi}}^{(t)})\}(\mathbf{A}\mathbf{A}^{T}+\varrho I_{r})^{-1}$

Normalize each column of $\mathbf{D_{1}}$ as $\mathbf{d_{1}}_{j}\leftarrow\frac{1}{\max(||\mathbf{d_{1}}_{j}||_{\infty},1)}\,\mathbf{d_{1}}_{j};$

Update $\mathbf{\mathbf{D_{2}}}$: for $1\leq j\leq p$, 
\begin{equation}
\mathbf{d_{2}}_{j}^{(t+1)}=\arg\min_{\mathbf{d_{2}}_{j}}\frac{\varrho}{2}||\mathbf{d_{2}}_{j}-(\mathbf{d_{1}}_{j}^{(t+1)}+\mathbf{\xi}_{j}^{(t)})||_{2}^{2}+\log\{1+\Psi_{\lambda_{1}}(\mathbf{d_{2}}_{j})\};
\end{equation}

$\mathbf{\xi}^{(t+1)}\leftarrow\mathbf{\xi}^{(k)}+(\mathbf{D_{1}}^{(k+1)}-\mathbf{D_{2}}^{(k+1)});$

$t=t+1$\; }

Remove the zero columns of $\mathbf{D_{2}}$;\\
 \Output{$\mathbf{D_{2}}$}

\protect\caption{\label{alg:Updic}Update dictionary using ADMM}
\end{algorithm}

\textbf{Details in implementation}. As demonstrated above, GSCAD has
the ability to prune dictionary, and later in Section $\ref{sub:Synthetic-Experiments}$,
we will see that its empirical performance is promising and competitive.
However, if we initiate dictionary with a size smaller than the truth,
there is nothing left for GSCAD to help. Therefore, an over-sized
dictionary in the initiation step is strongly preferred. Experiments
have shown that there is nothing to lose to start with a large dictionary
as GSCAD can always prune it to the right size.

During the dictionary updating stage after we obtain a new dictionary
from ADMM, if any two atoms are highly correlated, correlation greater
than 0.95 for example, we only keep one of them. Some experiments
have shown that this does not have much effect on the results, but
will speed up convergence of the algorithm.

We define the convergence of the algorithm by the differences of $\mathbf{D}$
and the differences of $\mathbf{A}$ between two consecutive iterations.
If they are both below a certain threshold, the algorithm stops. However,
in implementation, we add an extra rule on the maximum number of iterations,
since GSCAD may get stuck to a region where $\mathbf{D}$ keeps alternating
from two local minima and never converge due to a bad initiation.
Fortunately, the performance of local minima is mostly decent in terms
of denoising.

\section{Experimental Results}

\subsection{Synthetic Experiments\label{sub:Synthetic-Experiments}}

We design a simple example to check the performance of GSCAD from
two aspects: (i) whether GSCAD could recover the true size of the
dictionary, and (ii) its denoising performance compared with other
methods.

\begin{figure}
\begin{centering}
\includegraphics[width=14cm]{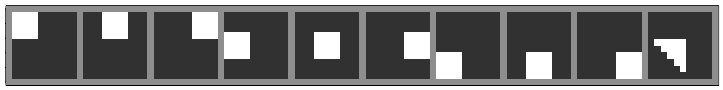} 
\par\end{centering}

\protect\caption{\label{fig:D0}Atoms of the generating dictionary $\mathbf{D}_{0}$.
Each atom corresponds to a $10\times10$ patch with white region representing
1 and black region representing 0.}
\end{figure}

\textbf{Data generation}. The generating dictionary $\mathbf{D}_{0}\in\mathbb{R}^{10\times100}$
contains 10 atoms. Each atom is a vectoried $10\times10$ patch shown
in Figure $\ref{fig:D0}$. Then 1500 signals $\{\mathbf{y}_{i}\}_{i=1}^{1500}$
in $\mathbb{R}^{100}$ are generated, each created by a linear combination
of three different generating dictionary atoms picked randomly, with
identically independently distributed coefficients following $Unif(0,1/3)$.
Gaussian noises $\epsilon_{i}\sim\mathcal{N}(0,\sigma^{2})$ are added,
with signal-to-noise ratio (SNR) controlled by the Gaussian variance
$\sigma^{2}$. Four levels of noise level are adopted at $\sigma=0.01,0.025,0.05,0.1$.

\textbf{Applying GSCAD}. In order to examine GSCAD's ability to prune
dictionaries to the right size, dictionaries are initialized with
varying number of atoms $p_{0}$, namely, 10 (true size), 15, 20 and
50. We run the GSCAD and received a learned dictionary $\hat{\mathbf{D}}\in\mathbb{R}^{m\times\hat{p}}$,
where the resulting size of the dictionary $\hat{p}$ might be, and
in most of the cases, is less than the initial size $p_{0}$. For
validation, another 1000 signals are generated under the same setting.
Both clean signals $\{\mathbf{x}_{i}^{test}\}_{i=1}^{1000}$ and noisy
signals $\{\mathbf{y}_{i}^{test}\}_{i=1}^{1000}$ are recorded. Coefficients
$\hat{\alpha}_{i}^{test}\in\mathbb{R}^{\hat{p}}$ corresponding to
$\mathbf{y}_{i}^{test}$ are obtained using Orthogonal Matching Pursuit(OMP)
with the number of non-zero elements fixed to three. We then reconstruct
signals as $\hat{\mathbf{x}}_{i}^{test}=\mathbf{\hat{D}}\hat{\alpha}_{i}^{test}$,
and calculate the PSNR as 
\[
\mbox{PSNR}=10\log_{10}(\frac{\sum_{i}||\mathbf{x}_{i}^{test}||^{2}}{\sum_{i}||\hat{\mathbf{x}}_{i}^{test}-\mathbf{x}_{i}^{test}||^{2}}).
\]

\textbf{Comparison}. For each setting, we also run the K-SVD algorithm
using the Matlab Toolbox associated its original paper \citet{aharon2006},
and Online Learning algorithm \citep{mairal2010online} using the
SPAMS package. Since neither K-SVD nor Online Learning would prune
the dictionary, the learned dictionary $\hat{\mathbf{D}}$ will be
in the same space as its initial value $\hat{\mathbf{D}}_{0}$, i.e.
$\hat{p}=p_{0}$. Validation for both method are conducted in the
same fashion as that in GSCAD.

\begin{figure}
\includegraphics[width=14cm]{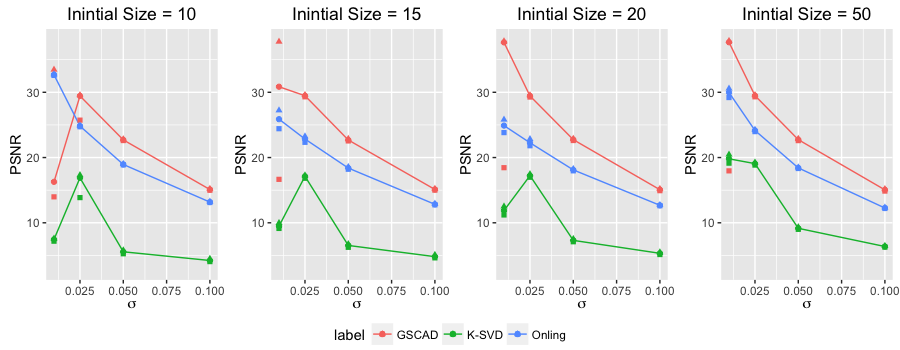}

\protect\caption{\label{fig:PSNR} Synthetic results. Median PSNR: circle points connected
by lines; first quartile: triangle-shaped points; third quartile:
square-shaped points. Different algorithms are indexed by color.}
\end{figure}

\textbf{Result}. For each setting of $\sigma$ and $p_{0}$, experiments
using GSCAD, Online Learning and K-SVD are repeated 100 times. Median,
first quartile and third quartile of the PSNR are shown in Figure
$\ref{fig:PSNR}$. GSCAD performs better consistently than the other
two methods when varying initial size $p_{0}$ and SNR levels controlled
by $\sigma$, except just one case when initial $p_{0}$ is at the
true value 10 and $\sigma=0.01$. As suggested in Section $\ref{sub:Algorithms}$,
to make fully use of GSCAD's power of pruning, it is better to start
with an over-sized initial dictionary. The mean and standard deviation
of $\hat{p}$, the size of the resulting dictionary for GSCAD are
reported in Table $\ref{tab:natom}$. The resulting size of the dictionary
learned from GSCAD are very close to the truth, with very small standard
deviations across all cases.

\begin{table}
\begin{centering}
\begin{tabular}{ccccc}
\hline 
$p_{0}\backslash\sigma$  & 0.01  & 0.025  & 0.05  & 0.1\tabularnewline
\hline 
10  & 9.97 (0.171)  & 10.00 (0)  & 10.00 (0)  & 10.00 (0)\tabularnewline
15  & 10.02 (0.245)  & 10.03 (0.171)  & 10.15 (0.359)  & 10.14 (0.403)\tabularnewline
20  & 10.02 (0.245)  & 10.08 (0.273)  & 10.15 (0.359)  & 10.35 (0.626)\tabularnewline
50  & 10.07 (0.293)  & 10.07 (0.293)  & 10.19 (0.394)  & 10.36 (0.644)\tabularnewline
\hline 
\end{tabular}
\par\end{centering}

\protect\protect\caption{\label{tab:natom}Average number of atoms in the resulting dictionary.
Numbers in the parenthesis are corresponding standard deviations.}
\end{table}

\subsection{Image Denoising}

\begin{figure}
\begin{centering}
\begin{tabular}{cc}
\includegraphics[width=5cm]{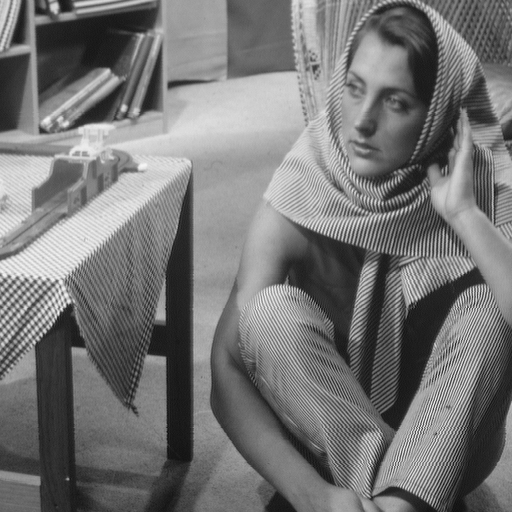} & \includegraphics[width=5cm]{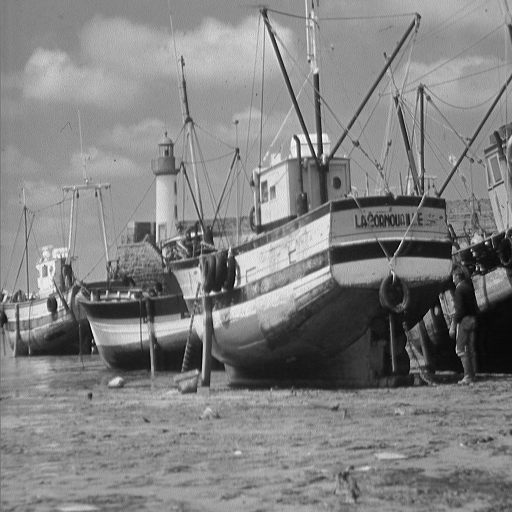} \tabularnewline
\includegraphics[width=5cm]{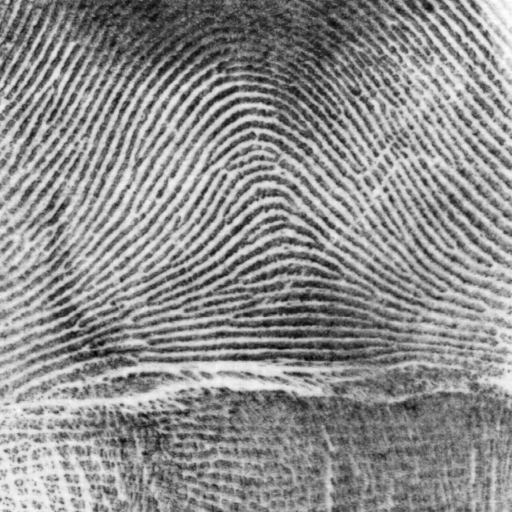} & \includegraphics[width=5cm]{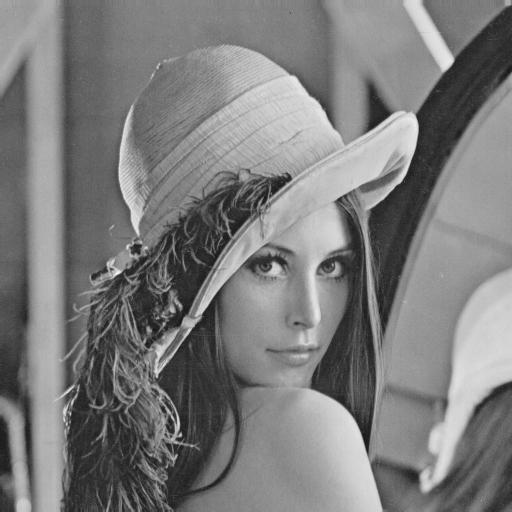} \tabularnewline
\includegraphics[width=5cm]{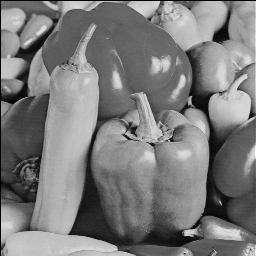} & \tabularnewline
\end{tabular}
\par\end{centering}

\caption{\label{fig:Images}Images used in denoising: Barb, Boat, Fgprt, Lena,
Peppers.}
\end{figure}

Following the denoising scheme proposed by \citep{elad2006}, we train
our dictionaries directly on patches from corrupted images. More details
about the scheme can be found in \citep{Mairal2014}. Five classical
images (\ref{fig:Images}) used in the image denoising benchmarks
are corrupted with Gaussian noise. Standard deviations of Gaussian
noise are set to be $\{5,10,20,50\}$ separately, for pixel values
in the range $[0,255]$. For each corrupted image, overlapped $8\times8$
patches are obtained and centered as training set. For an image of
size $512\times512$, a total number of 255025 patches $\mathbf{y}_{i}^{c}\in\mathbb{R}^{64}$
are extracted from the original image.

Dictionaries are trained using (i) the proposed GSCAD algorithm with
$\lambda_{1}=0.05$ and $c=3$ for Algorithm $\ref{alg:Updic}$, (ii)
K-SVD algorithm from the KSVD Matlab Toolbox, and (iii) Online Learning
algorithm from the SPAMS package. Redundant DCT of size size $p=256$
is used to initialize $\mathbf{D}$ for all three methods. The resulting
dictionary is in $\mathbb{R}^{64\times\hat{p}}$, where $\hat{p}$
stays the same as $p$ for K-SVD and Online Learning, but might be
smaller for GSCAD.

Once a dictionary $\hat{\mathbf{D}}$ is obtained, patches $\mathbf{y}_{i}^{c}$
are approximated up to the noise level by a sparse linear combination
of atoms in the dictionary: 
\[
\min_{\alpha_{i}\in\mathbb{R}^{\hat{p}}}||\alpha_{i}||_{0}\ s.t.\ ||\mathbf{y}_{i}^{c}-\hat{\mathbf{D}}\alpha_{i}||_{2}^{2}\leq\epsilon_{0},
\]
where $\epsilon_{0}$ is proportional to the noise variance $\sigma^{2}$.
We set $\epsilon_{0}=\sigma^{2}F_{m}^{-1}(\tau)$ with $\tau=0.9$
following the effective heuristic by \citep{mairal2009non}. $F_{m}$
is the $cdf.$ of $\chi^{2}$ distribution with freedom of $m$. Then
the denoised image is reconstructed based the sparse approximation
$\hat{\mathbf{y}}_{i}^{c}=\hat{\mathbf{D}}\hat{\alpha}_{i}$, and
its mean squared error ($MSE$) comparing to the clean image is calculated.
For each setting, the whole procedure is repeated five times with
different realizations of the noise. Define PSNR as 
\[
\mbox{PSNR}=10\log_{10}(255^{2}/\mbox{MSE}).
\]

The results for all three methods are very close to each other in
general. At lower noise levels, GSCAD has a better performance. 
$\ref{tab:natom dn}$ summarizes the average number of atoms for the
resulting dictionaries from GSCAD. It shows that GSCAD outperforms
the other two methods with a learned dictionary less than half of
the size of that used by the other algorithms. On the other hand,
at higher noise levels, GSCAD becomes less competitive. Specifically,
when $\sigma=50$, all resulting $\hat{p}$ are close to the initial
value of 256, which may indicate that dictionaries of size 256 is
not large enough as an initial value for GSCAD. Our experience suggests
that the higher noise level requires the larger dictionary size. Another
interesting finding is that bearing the same level of noise, image
'fingerprint' needs a much larger dictionary to denoise compared with
other images.


\begin{table}
\begin{centering}
\begin{tabular}{|c|ccc|ccc|ccc|}
\hline 
 & \multicolumn{3}{c|}{Barb} & \multicolumn{3}{c|}{Boat} & \multicolumn{3}{c|}{Fgrpt}\tabularnewline
\hline 
\hline 
$\sigma$  & Gscad  & Ksvd  & Online  & Gscad  & Ksvd  & Online  & Gscad  & Ksvd  & Online\tabularnewline
\hline 
5  & 38.23  & 38.05  & 38.15  & 37.24  & 37.24  & 37.06  & 36.60  & 36.65  & 36.53\tabularnewline
\hline 
10  & 34.57  & 34.39  & 34.73  & 33.70  & 33.64  & 33.81  & 32.38  & 32.44  & 32.52\tabularnewline
\hline 
20  & 30.79  & 30.90  & 31.13  & 30.30  & 30.46  & 30.63  & 28.31  & 28.58  & 28.70\tabularnewline
\hline 
50  & 25.51  & 25.75  & 25.87  & 25.79  & 26.09  & 26.25  & 23.00  & 23.63  & 23.59\tabularnewline
\hline 
\end{tabular}
\par\end{centering}

\begin{centering}
\begin{tabular}{|c|ccc|ccc|ccc|}
\hline 
 & \multicolumn{3}{c|}{Lena} & \multicolumn{3}{c|}{Peppers} & \multicolumn{3}{c|}{Average}\tabularnewline
\hline 
\hline 
$\sigma$  & Gscad  & Ksvd  & Online  & Gscad  & Ksvd  & Online  & Gscad  & Ksvd  & Online\tabularnewline
\hline 
5  & 38.63  & 38.60  & 38.56  & 38.04  & 37.78  & 37.81  & 37.75  & 37.66  & 37.62\tabularnewline
\hline 
10  & 35.58  & 35.48  & 35.70  & 34.51  & 34.22  & 34.71  & 34.15  & 34.03  & 34.29\tabularnewline
\hline 
20  & 32.31  & 32.41  & 32.60  & 30.84  & 30.82  & 31.22  & 30.51  & 30.63  & 30.86\tabularnewline
\hline 
50  & 27.66  & 27.88  & 28.02  & 25.84  & 26.32  & 26.37  & 25.56  & 25.93  & 26.02\tabularnewline
\hline 
\end{tabular}
\par\end{centering}

\protect\protect\caption{\label{tab:Averge-PSNR}Average PSNR over five runs.}
\end{table}

\begin{table}
\centering{}%
\begin{tabular}{cccccc}
\hline 
$\sigma$  & Barb  & Boat  & Fgrpt  & Lena  & Peppers\tabularnewline
\hline 
\hline 
5  & 109  & 125  & 178  & 86  & 100\tabularnewline
\hline 
10  & 92  & 94  & 195  & 80  & 86\tabularnewline
\hline 
20  & 120  & 151  & 224  & 136  & 153\tabularnewline
\hline 
50  & 248  & 248  & 246  & 250  & 247\tabularnewline
\hline 
\end{tabular}\protect\protect\caption{\label{tab:natom dn}Number of atoms for the resulting dictionary
($\hat{p}$).}
\end{table}

\section{Conclusion}

The GSCAD method has been presented for learning a sparse dictionary
and selecting the dictionary size simultaneously. The experimental
analysis has demonstrated very encouraging results relative to the
state-of-the-art methods. This new framework may also be applied to
the general subspace clustering problem for imaging clustering, which
assumes that similar points are described as points lying in the same
subspace. The proposed formulation can learn the clustering and the
number of clusters at the same time. This framework may also be applied
to the architecture design of deep learning. The new GSCAD penalty
can learn a sparse connection between units of two layers in the deep
neural network to improve efficiency.

\section*{Appendix}

\textbf{Proof of Theorem 1.}

(1). When $z_{k}=0$, we have $(z_{k}-0)^{2}\leq(z_{k}-\theta_{k})^{2}$,
and further 
\[
\log\{1+\psi_{\lambda}(0)+\sum_{l\neq k}\psi_{\lambda}(\theta_{l})\}\leq\log\{1+\psi_{\lambda}(\theta_{k})+\sum_{l\neq k}\psi_{\lambda}(\theta_{l})\},
\]
for any $\theta_{k}\in\mathbb{R}$. When $z_{k}\neq0$, we have 
\[
\{z_{k}-sign(z_{k})|\theta_{k}|\}^{2}\leq[z_{k}-\{-sign(z_{k})|\theta_{k}|\}]{}^{2},
\]
and further 
\[
\log\{1+\psi_{\lambda}(sign(z_{k})|\theta_{k}|)+\sum_{l\neq k}\psi_{\lambda}(\theta_{l})\}=\log\{1+\psi_{\lambda}(-sign(z_{k})|\theta_{k}|)+\sum_{l\neq k}\psi_{\lambda}(\theta_{l})\}.
\]
Therefore to minimize(2), $\hat{\theta}_{k}$ has to satisfy $sign(\hat{\theta}_{k})=sign(z_{k})$. If we denote $\tilde{K}=\{1\leq k\leq K:\ z_{k}\neq0\}$
and denote $\Theta_{k}$ as the open interval between $z_{k}$ and 0, i.e.
\[
\Theta_{k}=\begin{cases}
(0,z_{k}), & if\ z_{k}>0\\
(z_{k},0), & if\ z_{k}<0
\end{cases},
\]
then optimization problem (2) is equivalent to 
\[
\min_{\theta_{k}\in\Theta_{k}\cup\{0\},k\in\tilde{K}}\frac{\varrho}{2}\sum_{k\in\tilde{K}}(z_{k}-\theta_{k})^{2}+\log\{1+\sum_{k\in\tilde{K}}\psi_{\lambda}(\theta_{k})\}.
\]

(2). Recall that $c_{0}=card(\tilde{K})$. To simplify the notation, we rewrite $z=(z_{1},...,z_{c_{0}})\in\mathbb{R}^{c_{0}}$ as if there were no zero-element in $z$,
and correspondingly $\theta=(\theta_{1},...,\theta_{c_{0}})\in\mathbb{R}^{c_{0}}$.
Define $L:\mathbb{R}^{c_{0}}\rightarrow\mathbb{R}$:
\[
L(\theta)=\frac{\varrho}{2}||z_{k}-\theta_{k}||^{2}+\log\{1+\sum_{k=1}^{c_{0}}\psi_{\lambda}(\theta_{k})\}.
\]
We expend $\Theta_{k}$ to the whole half plane:
\[
\tilde{\Theta}_{k}=\begin{cases}
(0,\infty), & if\ z_{k}>0\\
(-\infty,0), & if\ z_{k}<0
\end{cases}.
\]
If we can show that $L$ is convex in $\tilde{\Theta}_{1}\times...\times\tilde{\Theta}_{c_{0}}$,
this will inply that $L$ is convex over $\prod_{k=1}^{c_{0}}\Theta_{k}\cup\{0\}$,
as $L$ is continous all over $\mathbb{R}^{c_{0}}$.

To show that the optimization problem within $\Theta^{o}=\tilde{\Theta}_{1}\times...\times\tilde{\Theta}_{c_{0}}$
is convex, we are going to varify the inequality 
\[
L((1-t)x+ty)\leq(1-t)L(x)+tL(y),\quad t\in[0,1],
\]
for any $x,y\in\Theta^{o}$. This is trivial for $x=y$, and for $x\neq y$,
we consider the following two cases.

Case 1: $x,y\in\Theta_{1}^{o}=\{x\in\Theta^{o}:|x_{i}|\notin\{\lambda,c\lambda\}\text{ for any }1\leq i\leq c_{0}\}$.
Therefore only a finite number of points in set $\{tx+(1-t)y:t\in[0,1]\}$
such that $L$ does not have a second-order derivative. Let $v=x-y$.
Define $\varphi(t)=L(x+tv),t\in[0,1]$. If we can show that $\varphi'(t)$
is continuous on $[0,1]$, and $\varphi''(t)\geq0$ except at a finite
number of points, therefore $\varphi'(t)$ is non-decreasing, and furthermore
$\varphi(t)$ is convex on $[0,1]$. By definition, for any $t\in[0,1]$,
\[
L((1-t)x+ty)=L(x+tv)=\varphi(t)\leq t\varphi(1)+(1-t)\varphi(0)=tL(y)+(1-t)L(x).
\]
Therefore $L$ is convex.

Now we are going to show that $\varphi'(t)$ is continuous and $\varphi''(t)\geq0$
except at a finite number of points, where $\varphi''(t)$ does not
exist. Taking derivative of $L$, we get 
\begin{align*}
L'_{x_{i}} & =sign(x_{i})\{\varrho|x_{i}|+\frac{\dot{\psi}_{\lambda}(x_{i})}{1+\sum_{k}\psi_{\lambda}(x_{k})}\}-\varrho z_{k},\\
L''_{x_{i}x_{i}} & =\varrho+\frac{\ddot{\psi}_{\lambda}(x_{i})}{1+\sum_{k}\psi_{\lambda}(x_{k})}-\frac{\dot{\psi}_{\lambda}^{2}(x_{i})}{\{1+\sum_{k}\psi_{\lambda}(x_{k})\}^{2}},\quad|x_{i}|\notin\{\lambda,c\lambda\},\\
L''_{x_{i}x_{j}} & =-\frac{\dot{\psi}_{\lambda}(x_{i})\cdot\dot{\psi}_{\lambda}(x_{j})}{\{1+\sum_{k}\psi_{\lambda}(x_{k})\}^{2}},\quad|x_{i}|,|x_{j}|\notin\{\lambda,c\lambda\}
\end{align*}
where 
\[
\dot{\psi}_{\lambda}(x_{i})=\begin{cases}
\lambda\cdot sign(x_{i}), & if\ |x_{i}|\leq\lambda\\
\frac{c\lambda-|x_{i}|}{(c-1)}\cdot sign(x_{i}), & if\ \lambda<|x_{i}|\leq c\lambda\\
0, & if\ |x_{i}|>c\lambda
\end{cases}\quad and\quad\ddot{\psi}_{\lambda}(x_{i})=\begin{cases}
-\frac{1}{(c-1)}, & if\ \lambda<|x_{i}|\leq c\lambda\\
0, & o.w.
\end{cases}.
\]
Since $L'_{x_{i}}$ is continuous for all $1\leq i\leq c_{0}$ and
$x\in\Theta^{o}$, 
\[
\varphi'(t)=\sum_{i}\frac{\partial L}{\partial x_{i}}(x+tv)\cdot v_{i}
\]
is contious. Except a finite number of $t\in[0,1]$, such that $L''_{x_{i}x_{j}}$
does not exist at $x+tv$, we have 
\begin{align*}
\varphi''(t) & =\sum_{i,j}\frac{\partial^{2}L}{\partial x_{i}\partial x_{j}}(x+tv)v_{i}v_{j}\\
 & =\sum_{i=1}^{c_{0}}\{\varrho+\frac{\ddot{\psi}_{\lambda}(x_{i})}{1+\sum_{k}\psi_{\lambda}(x_{k})}\}v_{i}^{2}-\{1+\sum_{k}\psi_{\lambda}(x_{k})\}^{-2}\{\sum_{i=1}^{c_{0}}\dot{\psi}_{\lambda}(x_{i})v_{i}\}^{2}\\
 & \geq\sum_{i=1}^{c_{0}}\{\varrho+\frac{\ddot{\psi}_{\lambda}(x_{i})}{1+\sum_{k}\psi_{\lambda}(x_{k})}\}v_{i}^{2}-\{1+\sum_{k}\psi_{\lambda}(x_{k})\}^{-2}c_{0}\sum_{i=1}^{c_{0}}\dot{\psi}_{\lambda}^{2}(x_{i})v_{i}^{2}\\
 & =\sum_{i=1}^{c_{0}}\{\varrho+\frac{\ddot{\psi}_{\lambda}(x_{i})}{1+\sum_{k}\psi_{\lambda}(x_{k})}-\frac{c_{0}\dot{\psi}_{\lambda}^{2}(x_{i})}{\{1+\sum_{k}\psi_{\lambda}(x_{k})\}^{2}}\}v_{i}^{2}.
\end{align*}
Let 
\[
f_{i}(x_{i})=\varrho+\frac{\ddot{\psi}_{\lambda}(x_{i})}{1+\sum_{l}\psi_{\lambda}(b_{l})}-\frac{c_{0}\,\dot{\psi}_{\lambda}^{2}(x_{i})}{\{1+\sum_{l}\psi_{\lambda}(b_{l})\}^{2}},\quad1\leq i\leq c_{0}.
\]
To show that $\varphi''(t)\geq0$, we only need to show that $f_{i}(x_{i})\geq0$.
Since $f_{i}(x_{i})=f_{i}(-x_{i})$, without loss of generality, we
are only going to show that $f_{i}(x_{i})\geq0$, for $x_{i}>0$.

Take derivative of $f_{i}$, 
\[
f_{i}'(x_{i})=-\frac{\ddot{\psi}_{\lambda}(x_{i})\dot{\psi}_{\lambda}(x_{i})}{1+\sum_{l}\psi_{\lambda}(x_{l})}-\frac{2c_{0}\,\dot{\psi}_{\lambda}^{2}(x_{i})\ddot{\psi}_{\lambda}(x_{i})}{\{1+\sum_{l}\psi_{\lambda}(x_{l})\}^{2}}+\frac{2c_{0}\,\dot{\psi}_{\lambda}^{3}(x_{i})}{\{1+\sum_{l}\psi_{\lambda}(x_{l})\}^{3}},\quad x_{i}\notin\{\lambda,c\lambda\}.
\]
Since $\ddot{\psi}_{\lambda}(x_{i})\leq0$ and $\dot{\psi}_{\lambda}(x_{i})\geq0$,
we have $f_{i}'(x_{i})\geq0$ for all $x_{i}\in\tilde{\Theta}_{k}\backslash\{\lambda,c\lambda\}$.
Observe that $f_{i}(x_{i})$ is piece-wise continous on $(0,\lambda),(\lambda,c\lambda)$, and $(c\lambda, \infty)$. For $x_{i}\in(0,\lambda)$,
\[
f_{i}(x_{i})\geq\lim_{x_{i}\rightarrow0^{+}}f_{i}(x_{i})=\varrho-\frac{c_{0}\lambda^{2}}{\{1+\sum_{l\in\tilde{K},l\neq k}p_{\lambda}(x_{l})\}^{2}}\geq\varrho-c_{0}\lambda^{2}\geq0.
\]
For $x_{i}\in(\lambda,c\lambda)$ 
\begin{align*}
f_{i}(x_{i}) & \geq\lim_{x_{i}\rightarrow\lambda^{+}}f_{i}(x_{i})\\
 & =\varrho-\frac{1}{(c-1)\{1+\lambda^{2}+\sum_{l\neq k}\psi_{\lambda}(x_{l})\}}-\frac{c_{0}\lambda^{2}}{\{1+\lambda^{2}+\sum_{l\neq k}\psi_{\lambda}(x_{l})\}^{2}}\\
 & \geq\varrho-\frac{1}{(c-1)(1+\lambda^{2})}-\frac{c_{0}\lambda^{2}}{(1+\lambda^{2})^{2}}\\
 & =\frac{\varrho(c-1)(1+\lambda^{2})^{2}-(1+\lambda^{2})-c_{0}(c-1)\lambda^{2}}{(c-1)(1+\lambda^{2})^{2}}\\
 & \geq0.
\end{align*}
For $x_{i}\in(c\lambda,\infty)$, 
\begin{align*}
f_{i}(x_{i}) & \geq\lim_{x_{i}\rightarrow c\lambda^{+}}f_{i}(x_{i})=\varrho>0.
\end{align*}
Therefore $f_{i}(x_{i})\geq0$, for $x_{i}>0$, and furthermore, $\varphi''(t)\geq0$ except a finite number of $t\in[0,1]$. Thus we finished the proof of case 1.

Case 2: $x\in\Theta_{0}^{o}$ or $y\in\Theta_{0}^{o}$,
where $\Theta_{0}^{o}=\Theta^{o}\backslash\Theta_{1}^{o}=\{x\in\Theta^{o}:|x_{i}|=\lambda,\text{or }c\lambda\text{, for some }1\leq i\leq c_{0}\}$.
Without loss of generality, we assume that the last $c_{0}-k,\ 1\leq k\leq n$
elements of $x$ and $y$ are the same, and the rest are not, i.e.
$x_{i}\neq y_{i}$ for $1\leq i\leq k$ and $x_{i}=y_{i}$ for $k+1\leq i\leq c_{0}$.
Let $x^{*}=(x_{1},...,x_{k})$, $y^{*}=(y_{1},...,y_{k})$ and $v^{*}=y^{*}-x^{*}$.
Therefore only a finite number of $t\in[0,1]$ such that point $(1-t)x^{*}+ty^{*}$
belongs to $\mathcal{D}^{k}=\{x\in\tilde{\Theta}_{1}\times...\times\tilde{\Theta}_{k}:|x_{i}|=\lambda,\text{or }c\lambda\text{, for some }1\leq i\leq k\}$.

Let $w=(w_{1},...,w_{k})$, and define $g:\tilde{\Theta}_{i_{1}}\times...\times\tilde{\Theta}_{i_{k}}\rightarrow\mathbb{R},$
as 
\[
g(w)=L((w,x_{k+1},...,x_{c_{0}})).
\]
Define $\varphi^{*}(t)=g(x^{*}+tv^{*}),t\in[0,1]$. Then similar to
Case 1, we can show that 
\[
\frac{d\varphi^{*}}{dt}=\sum_{i}\frac{\partial g}{\partial x_{i}^{*}}(x^{*}+tv^{*})\cdot v_{i}^{*}=\sum_{i=1}^{k}\frac{\partial L}{\partial x_{i}}\big((x^{*}+tv^{*},x_{k+1},...,x_{n})\big)\cdot v_{i}^{*}
\]
is contious, and 
\begin{align*}
\frac{d^{2}\varphi^{*}}{dt^{2}} & =\sum_{i,j}\frac{\partial^{2}g}{\partial x_{i}^{*}\partial x_{j}^{*}}(x^{*}+tv^{*})v_{i}^{*}v_{j}^{*}\\
 & =\sum_{i,j=1}^{k}\frac{\partial^{2}L}{\partial x_{i}\partial x_{j}}\big((x^{*}+tv^{*},x_{k+1},...,x_{n})\big)v_{i}^{*}v_{j}^{*}\\
 & \geq0
\end{align*}
except a finite number of $t\in[0,1]$. Therefore $d\varphi^{*}/dt$
is non-decreasing, and further $\varphi^{*}(t)$ is convex on $[0,1]$.
By definition, for any $t\in[0,1]$, 
\begin{align*}
L((1-t)x+ty)=L(x+tv) & =g(x^{*}+tv^{*})\\
 & =\varphi^{*}(t)\leq t\varphi^{*}(1)+(1-t)\varphi^{*}(0)=tL(y)+(1-t)L(x).
\end{align*}

Thus, we finished the proof of case 2.

\bibliography{Archive}

\end{document}